\documentclass[preprint]{elsarticle}

\makeatletter
\def\ps@pprintTitle{%
 \let\@oddhead\@empty
 \let\@evenhead\@empty
 \def\@oddfoot{\reset@font\hfil\thepage\hfil}
 \let\@evenfoot\@oddfoot
}

\usepackage{comment}
\usepackage{caption}

\usepackage{graphics}
\usepackage{epsfig}
\usepackage{amsmath}
\usepackage{amsfonts}
\usepackage{cases}
\usepackage{amsthm}
\usepackage{{color}}
\usepackage{graphicx}
\usepackage{caption}
\usepackage{subcaption}
\usepackage{natbib}
\usepackage{placeins}
\usepackage{float}

\usepackage{multicol}
\usepackage{multirow} 
\usepackage{comment}
\catcode`@=11
\@addtoreset{equation}{section}
\renewcommand\theequation{\thesection.\@arabic\c@equation}
\catcode`@=12

\usepackage{geometry}
\usepackage{color}

\newcommand{\blue}[1]{{\color{blue} #1}}

\numberwithin{equation}{section}

\usepackage{color}

\usepackage{graphicx} 

\usepackage[table,xcdraw]{xcolor}

\usepackage{color}

\usepackage{graphicx} 
\usepackage{indentfirst}

\usepackage[plainpages=false]{hyperref}
\hypersetup{colorlinks=true,citecolor=blue,filecolor=black,linkcolor=red}

\begin{document}
	
	\begin{frontmatter}
    

         \title{ \bf Bridging Accuracy and Interpretability: Deep Learning with XAI for Breast Cancer Detection
         \tnoteref{mytitlenote}}

          \author[nitt]{Bishal Chhetri}
\ead{bishalc@iitk.ac.in}

\author[nitt]{B.V. Rathish Kumar}
\ead{kumarrathish0123@gmail.com}


\address[nitt]{Department of Mathematics and Statistics, Indian Institute of Technology Kanpur, Kanpur - 208016, Uttar Pradesh, India}

\begin{abstract}
 \begin{multicols}{2}
Breast cancer remains the most prevalent malignancy and a leading cause of cancer related mortality among women worldwide. Early detection is critical because it substantially improves survival outcomes.  In this study, we present an interpretable deep learning framework for the early detection of breast cancer using quantitative features extracted from digitized fine needle aspirate (FNA) images of breast masses. Our deep neural network, using ReLU activations, the Adam optimizer, and a binary cross-entropy loss, delivers state-of-the-art classification performance, achieving an accuracy of 0.992, precision of 1.000, recall of 0.977, and an F1 score of 0.988. These results substantially exceed the benchmarks reported in the literature. We evaluated the model under identical protocols against a suite of well-established algorithms (logistic regression, decision trees, random forests, stochastic gradient descent, K-nearest neighbors, and XGBoost) and found the deep model consistently superior on the same metrics. Recognizing that high predictive accuracy alone is insufficient for clinical adoption due to the black-box nature of deep learning models, we incorporated model-agnostic Explainable AI techniques such as SHAP and LIME to produce feature-level attributions and human-readable visualizations. These explanations quantify the contribution of each feature to individual predictions, support error analysis, and increase clinician trust, thus bridging the gap between performance and interpretability for real-world clinical use. The concave points feature of the cell nuclei is found to be the most influential feature positively impacting the classification task. This insight can be very helpful in improving the diagnosis and treatment of breast cancer by highlighting the key characteristics of breast tumors. Our study highlights the potential of deep learning models to enhance diagnostic capabilities and treatment planning for breast cancer, while highlighting the importance of interpretability in AI systems.

\end{multicols}

\end{abstract}
    
\end{frontmatter}

 \begin{multicols}{2}      
 \section{Introduction}   
 
Cancer is a worldwide epidemic that has become an major public health problem today. Cancer incidence is found to increase year by year. In a 2022 report by the Agency for Research on Cancer (IARC), 20 million new cases and deaths of around 9.7 million were reported globally, with lung and breast cancer being the most common \cite{example}. In India, around 1.39 million cancer cases were reported in 2020. This number increased to 1.42 million and 1.46 million in 2021 and 2022, respectively. Further forecasting suggests that there could be a 12.8$\%$ increase in the number of annual cases by 2025 \cite{sathishkumar2022cancer}. Cancer results from some changes at the genetic level that cause uncontrolled growth and division of cells \cite{roy2017cancer}. Accumulated abnormal cells invade the surrounding areas, or metastasize to other organs, and affect their functions \cite{chaffer2011perspective}. Most cells in the human body have specific functions and fixed lifespans. They undergo a natural process called apoptosis, in which they are programmed to die when no longer needed. This allows the body to replace old or damaged cells with new functional ones and maintain cell populations in tissues \cite{elmore2007apoptosis}. However, cancerous cells bypass these regulatory mechanisms. They stimulate some mechanisms and prevent cells from going through the apoptosis process, leading to uncontrolled growth and tumor formation \cite{loeb1938causes}.

There are various types of cancer, named based on their site of origin, and breast cancer is the most prevalent type. Breast cancer is a complex and multifaceted disease that has attracted significant attention in the medical and scientific communities \cite{harbeck2019breast}. According to \cite{BCstatistics}, in the US, approximately $1$ in $8$ women ($13\%$) are likely to get invasive breast cancer at some point in their life. Approximately $310,720$ women were diagnosed with invasive breast cancer in 2024 \cite{BCstatistics}. Breast cancer typically arises from benign ductal or lobular tissues. With some pioneering research breakthroughs, mutations in genes such as BRCA1 and BRCA2 have been found to increase the chances of having breast cancer \cite{mehrgou2016importance}.   Advances in knowledge of the human genome and the discovery of mutations have shown that cancer arises due to genetic and epigenetic alteration and dysregulation of molecular pathways \cite{magi2017current}. This understanding has paved the way for precision medicine, which aims to treat the disease based on the molecular profile of an individual. Early detection is the key to the successful treatment of almost all cancers.  Appropriate screening tools are crucial to detecting early signs of cancer. The most common screening techniques used for breast cancer include mammography, ultrasound, and thermography. These techniques are important screening tools, but can also become ineffective in some cases \cite{barba2021breast}.  Accurate interpretation of the images generated from imaging methods can be a significant challenge for even the best clinicians \cite{evans2012breast}. Mammography and other early detection methods can fail sometimes to identify early-stage tumors particularly in women with dense tissue \cite{barba2021breast}. Other major challenges identified are listed below.

\begin{enumerate}
    \item \textbf{Result Interpretation}: Proper interpretation of the images generated from imaging techniques such as mammography can be a significant challenge and can vary between medical professionals. This can lead to difference of opinion and differences in accuracy.
    \item \textbf{Risk in Manual Analysis}: There is always a risk in any manual analysis, and this can lead to missed or false negative cases.
    \item \textbf{Time Consuming and Costly}: Traditional approach of breast cancer detection is time consuming and financially demanding.
    \item \textbf{Limited Access}: Limited access to screening facilities especially in rural areas.
    \item \textbf{Lack of Awareness}: Lack of awareness about early cancer detection.
\end{enumerate}

\end{multicols}


\begin{multicols}{2} 
Artificial intelligence models have become powerful tools and are widely used for the early detection and diagnosis of various types of cancer, including breast, lung, and brain cancer. These models learn the complex patterns and relationships in the data and help physicians improve the detection rate, thereby leading to early treatment and better patient outcomes.  Numerous studies have shown the effectiveness of these techniques in analyzing medical images and predicting diseases. Deep learning models, in particular, have been shown to achieve high classification accuracy in detecting breast cancer using low-frequency electromagnetic fields and mammography \cite{akbari2023mammography}. These models have shown promising results in the analysis of various types of medical images, such as histopathological images, for the early detection. Artificial intelligence models have demonstrated outstanding performance in a wide range of fields. However, because of their complexity, they are often labeled as “black boxes.” In sectors such as military, healthcare, finance, etc., understanding the inner workings of the models and interpreting the predictions becomes important. To believe, and deploy the model, in these sectors, it is extremely important to understand how the algorithm has come to a particular decision. We need some tools or techniques to explain the predictions made by these models. This is the fundamental driving force behind explainable artificial intelligence (XAI) \cite{doshi2017towards, wani2024deepxplainer}.  XAI is a collection of techniques and methods that are used to explain the predictions made by AI models \cite{dwivedi2023explainable}. Interpretability techniques such as shapely additive explanations (SHAP) and local interpretable model-agnostic explanations (LIME) are widely used XAI methods today. SHAP is based on a game-theoretic approach. SHAP values are calculated for each features uisng Shapley values from co-operative game theory.  The main idea here is to fairly distribute total contribution among all the data features \cite{dwivedi2023explainable}.  SHAP values are calculated as average marginal contribution using the formula,
\[
\phi_i = \sum_{S \subseteq F \setminus \{i\}} \frac{|S|! (|F| - |S| - 1)!}{|F|!} \left( f(S \cup \{i\}) - f(S) \right)
\]
where $F$ is the total number of features, $S$ is the feature subsets without $i$, $f(S)$ denotes the model prediction using subsets, and $f(S \cup \{i\})$ is the model prediction including feature $i$. LIME, on the other hand, approximates any black box model with a local, interpretable model and explains each model prediction. More details about SHAP and LIME techniques can be found in  \cite{molnar2020interpretable, salih2024perspective}.

\vspace{.3cm}

Although several studies have explored early cancer detection using machine learning models, the integration of explainable AI (XAI) techniques such as SHAP values and LIME together remains limited.  In \cite{khater2023explainable}, the authors use SHAP and partial dependence plots to explain the model prediction but they do not use the LIME technique and moreover, the precision of their proposed model is lower. The authors in \cite{mohi2023xml, nayak2025explainable} provide explanations for their machine learning models, but they use only one XAI technique for feature ranking without exploring how the features specifically impact classification.  To address these gaps, we propose an explainable deep neural network model that uses ReLU activation, cross-entropy loss, and the Adam optimizer for early breast cancer detection.  To enhance the interpretability of the model predictions, both local and global explanation techniques such as SHAP and LIME are used. The organization of the paper is as follows.  In the immediate section, we discuss the dataset and data pre-processing techniques used for early breast cancer detection. In section 3, we briefly discuss the methodology used. The classifying performance of the neural network, logistic regression, decision tree, random forest, SGD, KNN, and XGBoost models are discussed in section 4 along with the model interpretation using SHAP and LIME methods. The discussion and conclusion are presented in section 5.

\section{Data and Data Pre-processing}
\subsection{\textbf{Dataset Used}}

\begin{figure}[hbt!]
\begin{center}
\includegraphics[width=7.6in, height= 2.0in, angle=0]{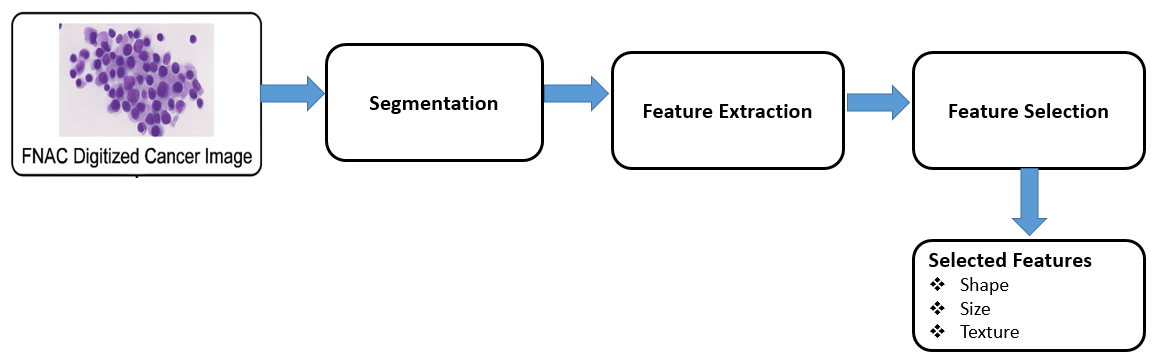}
\caption{Feature extraction workflow from a digitized cancer image. }
\label{feature_extraction}
\end{center}
\end{figure}

\begin{figure}[hbt!]
\begin{center}
\includegraphics[width=5.0in, height= 2.0in, angle=0]{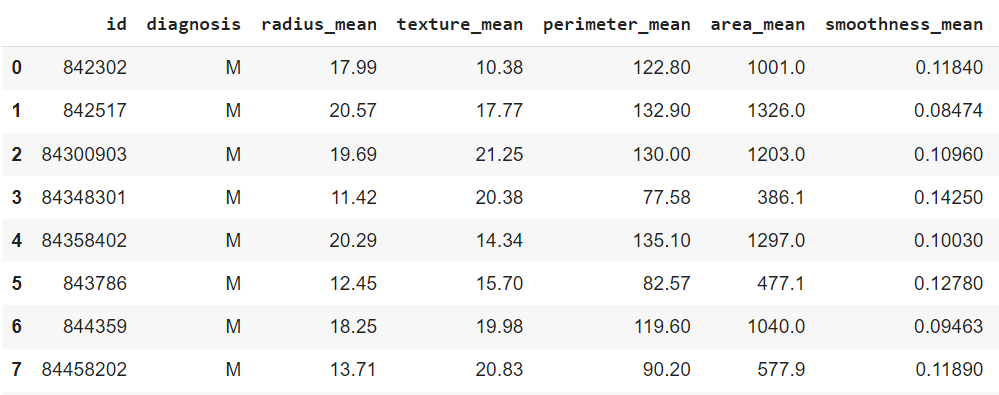}
\caption{First few columns of the original dataset.}
\label{datasetcol}
\end{center}
\end{figure}

We use the breast cancer dataset provided by the University of Wisconsin Research Institute \cite{breastcancerdata}. This dataset includes data from 569 patients, from whom features related to nuclear size, shape, and texture are extracted from digitized images of fine needle aspirates (FNA) taken from breast masses. Among these 569 patients, 357 were diagnosed with benign tumors, while the remaining patients were categorized as malignant (refer to Figure \ref{Distribution_M_and_B}). The size, shape, and texture-based features are used to differentiate between benign and malignant breast cytology. For digital analysis, images were generated from the stained glass slide using a JVC TK-1070U color video camera mounted on an Olympus microscope. The active contour model known as a "snakes" \cite{kass1988snakes} was used to accurately delineate the boundaries of each cell nucleus. The computer vision diagnostic system is designed to extract ten distinct features from the boundaries of cell nuclei generated by a snake algorithm. These features are radius (mean of the distance from center to points on the perimeter), smoothness (variation in radius lengths), perimeter, area, concavity (severity of concave portions of the contour), compactness, symmetry, texture (variation in gray-scale intensity), concave points (number of concave portions of the contour), and fractal dimension. For each image, the mean value, maximum value, and standard error for each feature are calculated, resulting in a total of 30 derived features. More detailed information on the data set and the feature extraction technique used can be found in \cite{street1993nuclear}. Figure \ref{feature_extraction} illustrate the workflow for feature extraction technique. The nuclear size, shape, and texture based features and their corresponding values are extracted from the images.  The numerical values of some of the extracted features can be viewed in Figure \ref{datasetcol}. The labels in the dataset (B or M) indicate whether each sample is classified as benign or malignant.

\begin{center}
\includegraphics[width=4.0in, height= 2.5in, angle=0]{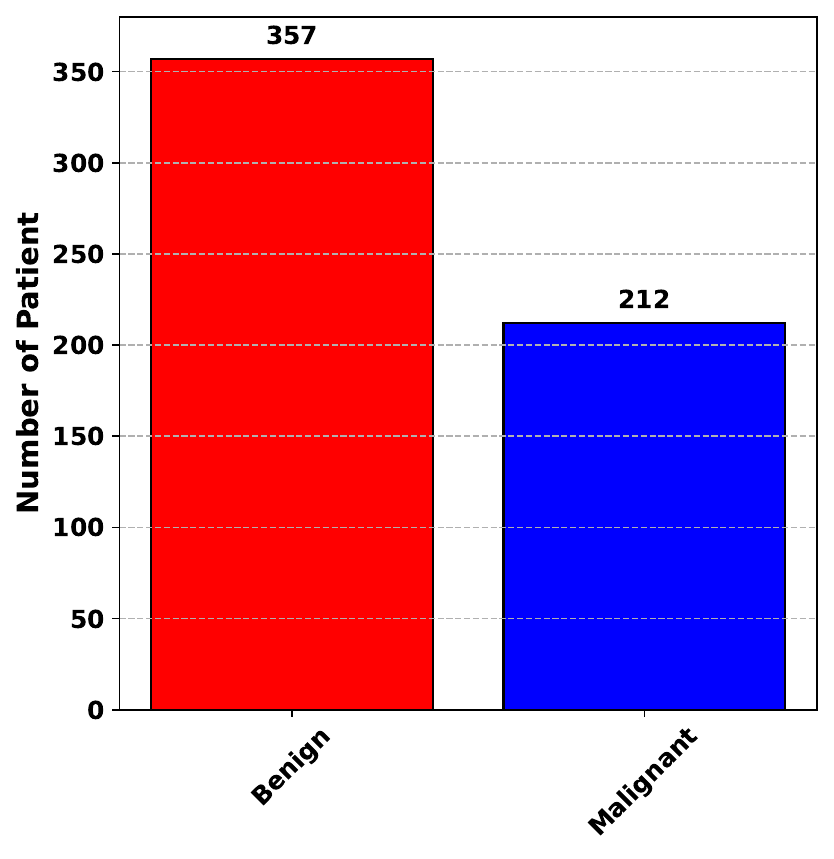}
\captionof{figure}{Number of benign and malignant cases in the dataset.}
\label{Distribution_M_and_B}
\end{center}

\end{multicols}

\begin{multicols}{2}
\subsection{\textbf{Data Scaling and Feature Selection}}
By analyzing the range of
values of each feature, we observe that the data values differ significantly (Figure \ref{datasetcol}). These differences may affect the performance of the models. To address this, we normalize the data using Min-Max scaling.
    \begin{equation}
    X_\text{normalized} = \frac{X - X_\text{min}}{X_\text{max} - X_\text{min}}  \label{norm1}
\end{equation}

Feature selection in a machine learning algorithm is an important part of the data pre-processing step. This step helps to find the optimal subset of features, reduces redundancy in the dataset, and improves the accuracy of the model. In our dataset, the feature named id was removed, as it serves merely as an identifier and has no predictive relationship with the target variable diagnosis. All other features were retained for model training and evaluation.

\section{Methodology Followed}
The goal here is to classify whether a patient has cancer or not. So, our problm is a binary classification problem. Amongst numerous classification algorithms, in this study, we use decision tree, logistic regression,   k-nearest neighbors,  random forest, stochastic gradient descent, XGBoost, and artificial neural network for classifying breast
cancer.  The raw numerical data obtained from the feature extraction process (Figure \ref{feature_extraction}) were pre-processed prior to training the model. Data were normalized on a scale of 0 to 1 using the Min-Max scaling function. The categorical variable, namely ``diagnosis'' is converted to numerical values 0/1 using LabelEncoder from sklearn. To solve the problem of sample imbalance, we use the stratified sampling method \cite{cao2022css}. This approach generates a more representative sample,  making it suitable for unbalanced samples. Data partitioning can also affect the performance of the models. The model's performance may be sensitive to data partitioning and may lead to different outputs. To check the sensitivities of these models,  we take two different cases. In the first, we divide the data into training and test sets in the ratio of 7:3, and in the second case, in the ratio of 8:2. The performance of the models is evaluated on the basis of precision, recall, accuracy, and F1 score. The accuracy gives the percentage of predictions that are correct and is defined by the formula;

\begin{align}
  \text{Accuracy} = \frac{\text{TN} + \text{TP}}{\text{TN} + \text{TP} + \text{FP} + \text{FN}}  
\end{align}

\begin{figure}[hbt!]  
\begin{center}
\includegraphics[width=6.2in, height= 3.0in, angle=0]{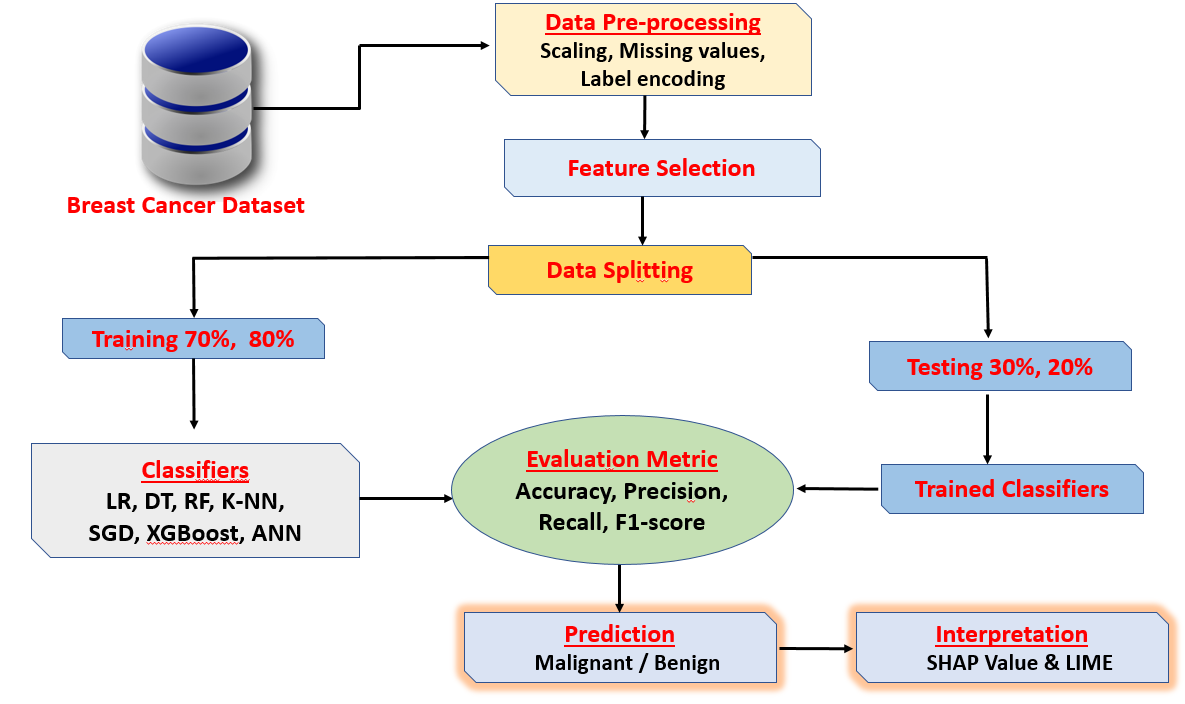}
\caption{Flowchart of the proposed methodology. The raw numerical data obtained from the feature extraction process were pre-processed and subsequently divided into training and testing sets prior to model training. }
\label{flowchart_ML}
\end{center}
\end{figure}

TP denotes the true positive case, it indicates the number of positive samples in the dataset that are correctly predicted as positive by the model (in our case, the number of malignant cases predicted malignant). The number of negative samples predicted as negative is denoted by TN.  The negative samples predicted as positive, and  positive samples predicted as negative are given by FP and FN respectively. The details of each of the evaluation metrics used are given in Table \ref{evaluation_metrics} and the flowchart of the proposed methodology is given in Figure \ref{flowchart_ML}. 

\end{multicols}

\begin{table}[H]
\caption{Evaluation Metrics for Binary Classification}
\centering
\begin{tabular}{|c|c|c|}
\hline
\textbf{Metric} & \textbf{Formula} & \textbf{Description} \\ 
\hline
Accuracy & $\frac{\text{TP} + \text{TN}}{\text{TP} + \text{TN} + \text{FP} + \text{FN}}$ & The percentage of predictions that are correct \\
\hline
Precision & $\frac{\text{TP}}{\text{TP} + \text{FP}}$ & The percentage of positive predictions that are correct \\
\hline
Sensitivity (Recall) & $\frac{\text{TP}}{\text{TP} + \text{FN}}$ & The percentage of positive cases that were predicted as positive \\ 
\hline
F1 Score & $ 2 \frac{\text{Precision $\times$ Recall}}{\text{Precision} + \text{Recall}}$ & F1 score balances Precision and Recall \\
\hline
\end{tabular} \label{evaluation_metrics}
\end{table}

\begin{multicols}{2}

\section{Results}
In the following, we evaluate the performance of various machine learning models based on accuracy, precision, recall, and F1-score and compare their performance with the performance of an artificial neural network model. 

\subsection{\textbf{Logistic Regression Prediction}}
The logistic regression model is one of the simplest models primarily used for binary classification problems. It uses a sigmoid function and produces a probability value between 0 and 1 and using a threshold value classification is made. The general form of logistic regression model is
\begin{equation*}
    y(x) = \frac{e^{\beta_0 + \beta_1 x}}{1 + e^{\beta_0 + \beta_1 x}  }
\end{equation*}
where $\beta_0$ and $\beta_1$ are regression coefficients and $x$ is the independent variable or input.\\

The dataset was initially divided in a 7:3 ratio for training and testing, and 5-fold cross-validation was performed. The mean accuracy from the cross-validation was recorded as 0.96734. After training the model, predictions were made on the test dataset, and the confusion matrix was calculated. The results showed the accuracy, precision, recall, and F1-score as 0.971, 1.0, 0.936, and 0.967, respectively. To assess the sensitivity of the model on the data split, we divided the entire data into training and test sets in 8:2 ratio. In this scenario, the mean cross-validation accuracy was found to be 0.9692, with accuracy, precision, recall, and F1-score recorded as 0.982, 1.0, 0.953, and 0.976, respectively. In this case, we find that the evaluation metric values almost remains the same.


\subsection{\textbf{Decision Tree Prediction}}
Decision tree, a type of supervised learning algorithm is widely used in regression and classification tasks. It has broad applications, the most common being in medical diagnosis, spam detection, risk assessment, and customer segmentation. The recursive partitioning of the data results in a tree-like structure. The algorithm selects the best attribute and splits the data at each internal node, based on certain criteria and the splitting process stops once a stopping criterion is met. The stopping criterion could be the maximum depth of the tree or having a minimum number of instances in a leaf node. \\

\begin{figure}[hbt!]
\begin{center}
\includegraphics[width=5.8in, height= 2.2in, angle=0]{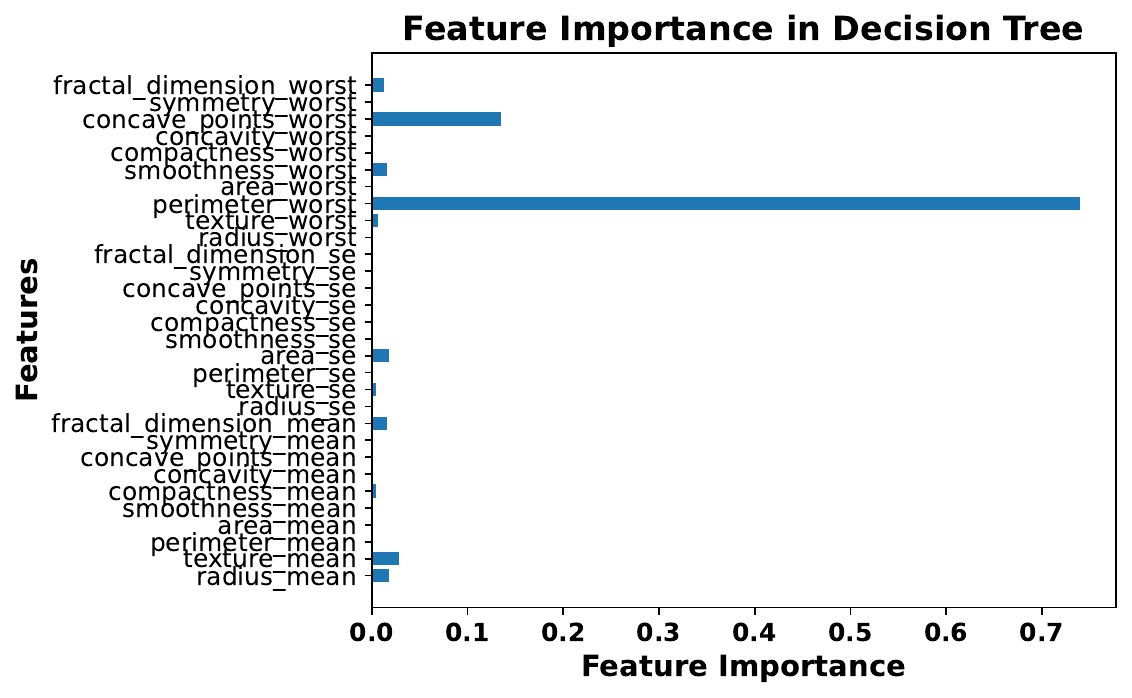}
\caption{Feature importance in decision tree classification. The decision tree classifier identifies perimeter$\_$worst as the most important feature for classifying cancer as benign or malignant.}
\label{DT_feature_impt}
\end{center}
\end{figure}

\begin{figure}[hbt!]
\begin{center}
\includegraphics[width=8.2in, height= 4.8in, angle=0]{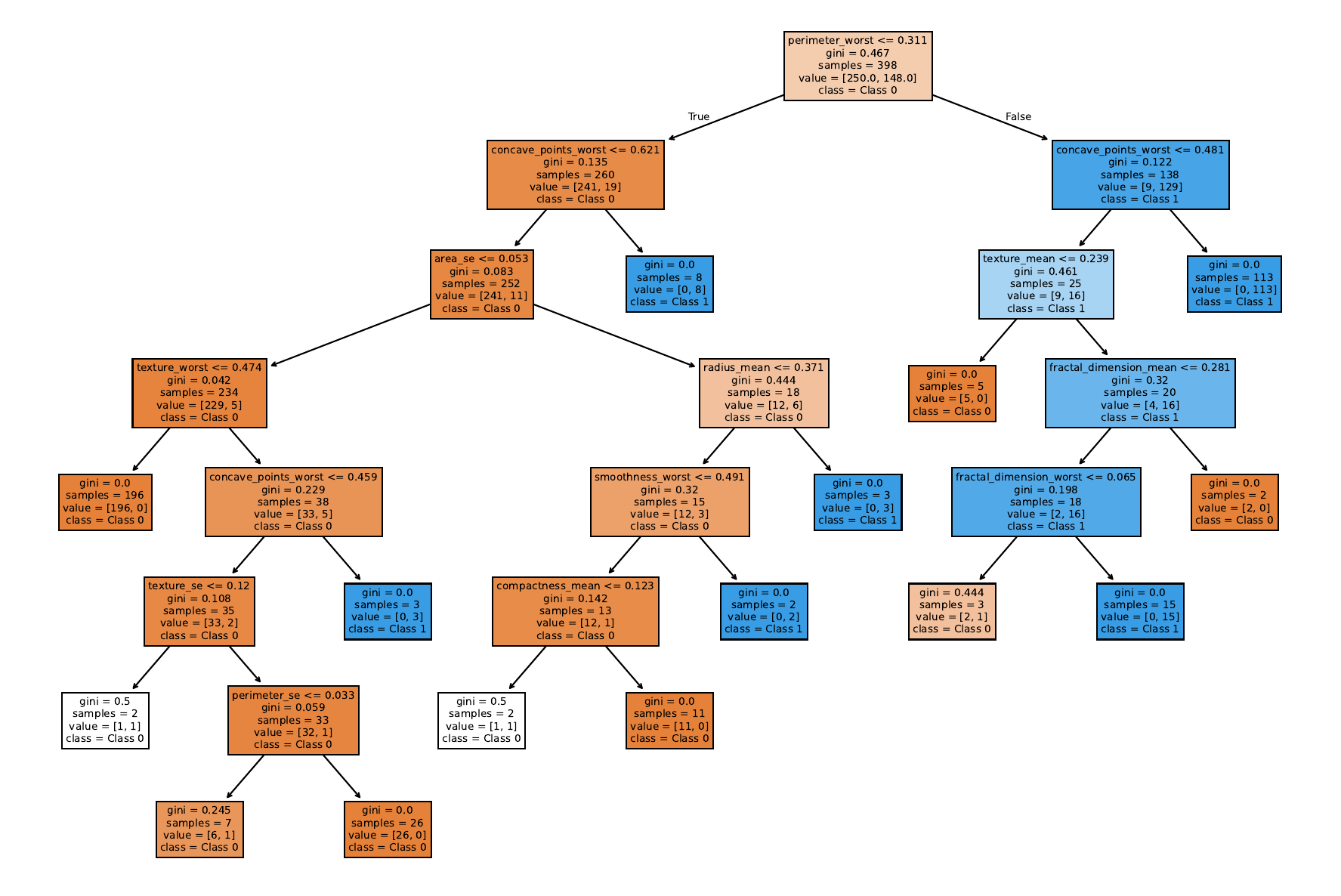}
\caption{Visualization of the Decision Tree. The feature perimeter$\_$worst is identified as the most important attribute for  splitting the data. }
\label{DT_visualization}
\end{center}
\end{figure}

All the attributes or the features in the data may not be so relevant in predicting the target variable. Feature importance in a decision tree refers to the contribution of a given feature in reducing the impurity (e.g., Gini impurity or entropy) or improving the model performance. In Figure \ref{DT_feature_impt}, the importance of each of the features is shown. The visualization of how decision tree splits the data and makes decisions is shown in Figure \ref{DT_visualization}.  We find that perimeter$\_$worst has a high importance level followed by concave$\_$points$\_$worst and smoothness$\_$worst. This indicates that these features play an important role in the model's predictions.  The accuracy, precision, recall, and F1-score for the 7 : 3 data split were  0.894, 0.835, 0.888, and 0.860 respectively. The mean cross-validation accuracy score was found to be 0.914.  For the 8:2 data split, the accuracy, precision, recall, and F1-score were found to be  0.901, 0.865, 0.90, and 0.882 respectively. We also performed hyperparameter tunning for decision tree using grid search. We explored different criteria ("gini" and "entropy"), maximum depths (10, 20, 50, 100, and 200), and minimum$\_$samples$\_$leaf (5, 10, 20, 50, and 100) to optimize model performance. The best parameters were {'criterion'= entropy, 'max  depth' = 10, 'min samples leaf'= 10} with a roc-auc score of 0.9583 (for 7:3 data split) and 0.9709 (for 8:2 data split). The accuracy, precision, recall, and F1-score for the tuned parameters for the two data split cases were    0.941, 0.895, 0.952, and 0.920 and 0.947, 0.930,  0.930, and 0.930  respectively. We find that the evaluation metric score increases with hyper-parameter tuning. The best scores for the evaluation metrics are shown in Table \ref{performance_comparison_combined}.

\subsection{\textbf{Random forest Prediction}}
Random forest is another powerful supervised learning algorithm used for both regression and classification. It employs multiple decision trees and makes predictions.  Random forests apply the general technique of bootstrap aggregating or bagging.  If we have a training set $X = x_1, ..., x_n$ and output $O = y_1, ..., y_n$, bagging repeatedly selects (say A times) a random sample with replacement from the training data and fits the decision trees to these samples \cite{rigatti2017random}. For classification tasks, random forest predicts the class selected by most trees and for regression, it is the average of the predictions \cite{ho1995random}.\\

The accuracy, precision, recall, and F1-score for the 7 : 3 data split were  0.970, 0.983, 0.936, and 0.960 respectively. The mean cross-validation accuracy was found to be 0.957.  For the 8:2 data split, the accuracy, precision, recall, and F1-score were found to be 0.956, 0.957, 0.930, and 0.941 respectively. Like decision tree, we also performed hyperparameter tunning using the grid search. For this we used max$\_$depth ( None, 10,20, 50, 100), number of estimators (10, 20, 30, 50, 100), and minimum sample leaf (1, 5, 10, 20, 50).  The best hyper-parameters identified were max$\_$depth = None, number of estimators = 30, and minimum sample leaf = 1. The roc-auc score in this case was found to be 0.99. Accuracy, precision, and F1-score increased in this case and were found to be 0.970, 0.983, and 0.960 respectively. 

\end{multicols}

\begin{multicols}{2}
\subsection{\textbf{KNN Prediction}}
KNN is yet another simple, but powerful, supervised machine learning algorithm that is widely being used in classification and regression tasks. For a given point or query, this algorithm finds the K nearest neighbors (K being a hyper-parameter is very important here) based on a distance metric. The prediction of the model is determined by the majority vote or the average of the K neighbors.  Euclidean distance is the most common distance metric used.\\

\textbf{\textit{Euclidean distance formula}}
\begin{equation*}
    d(x, y) = \sqrt{\sum_{i=1}^{n} (x_i - y_i)^2}
\end{equation*}
where $x$ and $y$ are the two points in an  n-dimensional space. To get an optimal value of K, we vary the value of K from 1 to 20. For each K, the five-fold cross-validation is performed and accuracy is calculated. Accuracy versus the value of K is plotted in Figure \ref{KNN_Accuracy}. We find that the accuracy is highest for K= 4.  Using the value of K = 4, we train the KNN model. For the 7:3 ratio data split, the accuracy, precision, recall, and F1-score were found to be 0.977,  0.984, 0.952, and  0.968 respectively.  For the 8: 2 ratio the values were 0.974,  0.976,  0.953, and  0.965. 

\end{multicols}

\begin{multicols}{2}
\subsection{\textbf{XGBoost}}
Due to its ability to handle large datasets,  Extreme Gradient Boosting is a popular choice in machine learning today. Because of its built-in support for parallel processing, large datasets can be trained in a relatively short amount of time. In XGBoost, a series of decision trees are constructed in sequence, with each new tree trying to rectify the mistakes of the preceding ones. The accuracy, precision, recall, and F1-score of the XGBoost model for two different cases of data split are shown in Table \ref{performance_comparison_combined}. We find that the XGBoost model has an accuracy of around 97$\%$ and shows better prediction performance when the data set is divided in the ratio of 7:3.  In this case, the XGBoost model is found to have better prediction accuracy compared to the Decision Tree and SGD algorithms.

\subsection{\textbf{SGD Algorithm}}
SGD is an iterative optimization process and it searches for the optimum value of the objective function. The basic idea here is to iteratively adjust parameters (like weights in a model) in the direction of the steepest decrease in the function's value \cite{geeks_sgd}.  Unlike other models, in SGD, for each iteration, a small batch of dataset is selected at random to calculate the gradient and update the model parameters reducing the computational cost significantly. The performance of the SGD algorithm on breast cancer classification is shown in Table \ref{performance_comparison_combined}. The model has an accuracy of 94$\%$. 

\begin{center}
\includegraphics[width=3.2in, height= 2.0in, angle=0]{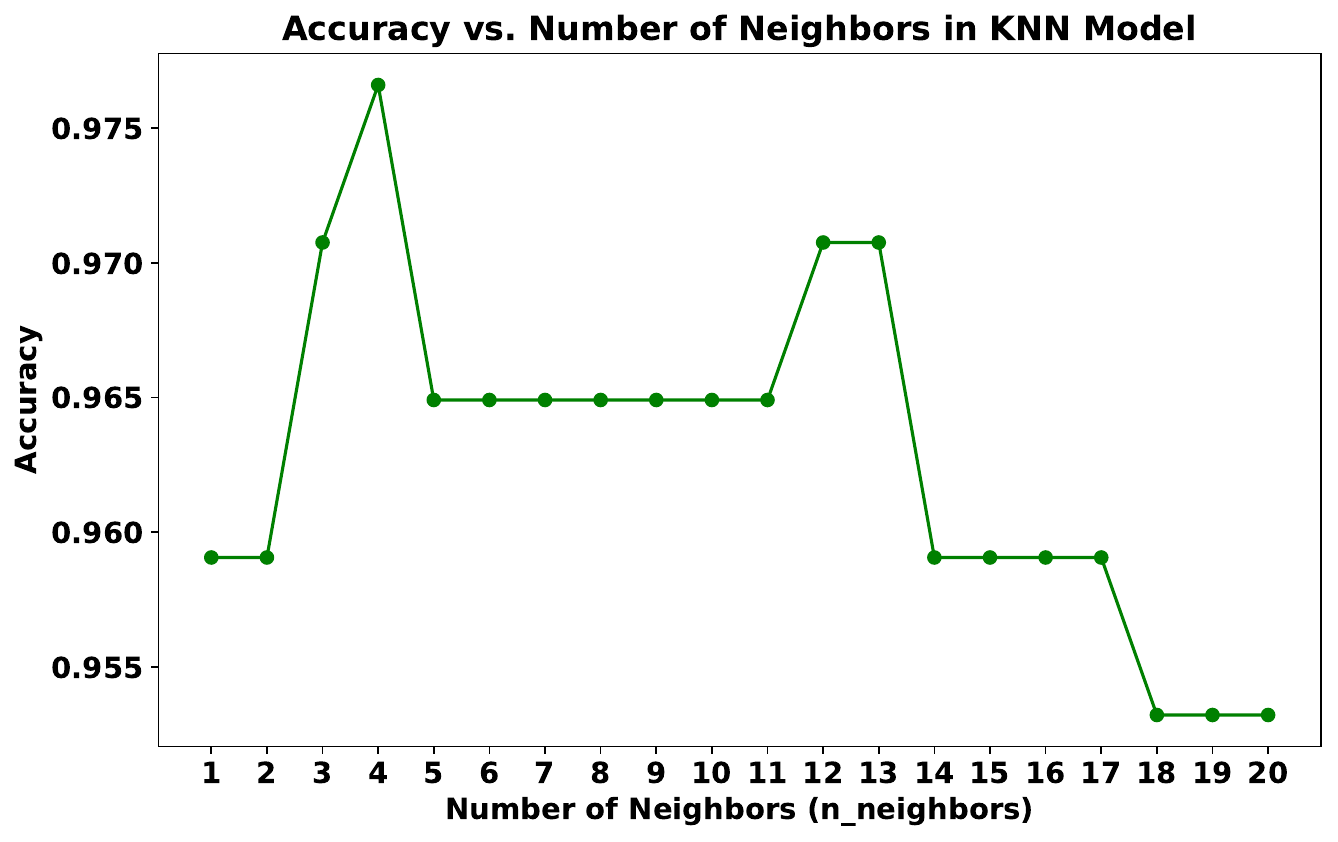}
\captionof{figure}{Accuracy of KNN model with different values of nearest neighbors (K).}
\label{KNN_Accuracy}
\end{center}

\subsection{\textbf{ANN Results}}
\begin{figure}[H]
\begin{center}
\includegraphics[width=3.8in, height= 2.0in, angle=0]{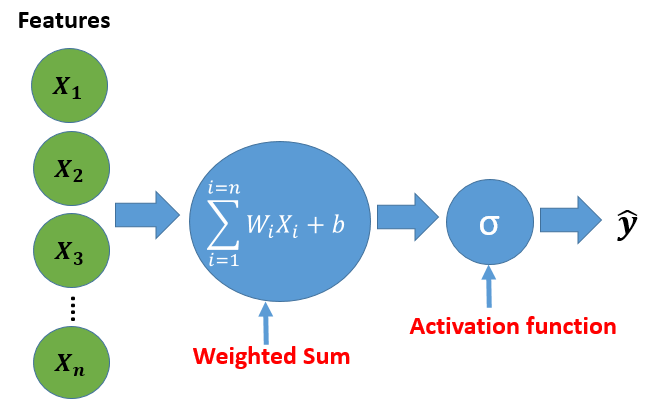}
\includegraphics[width=3.8in, height= 2.0in, angle=0]{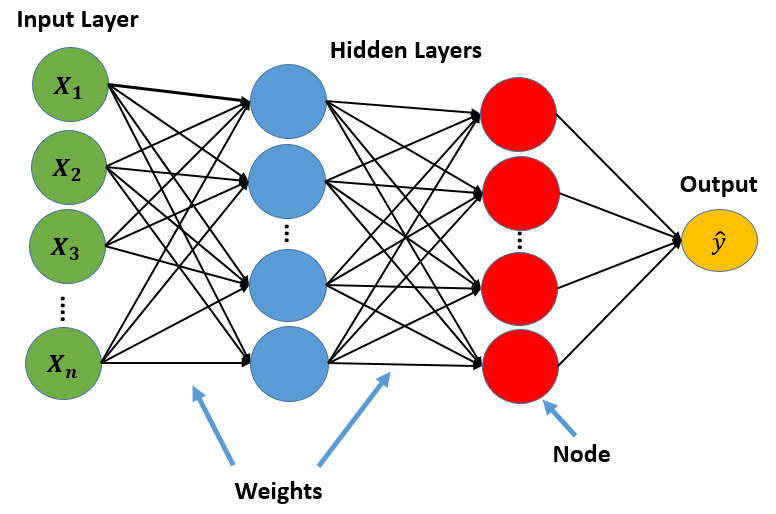}
\caption{Left) Perceptron. Right) Multi-layer Perceptron.}
\label{NN_figures}
\end{center}
\end{figure}

\begin{figure}[hbt!]
\begin{center}
\includegraphics[width=5.8in, height= 2.2in, angle=0]{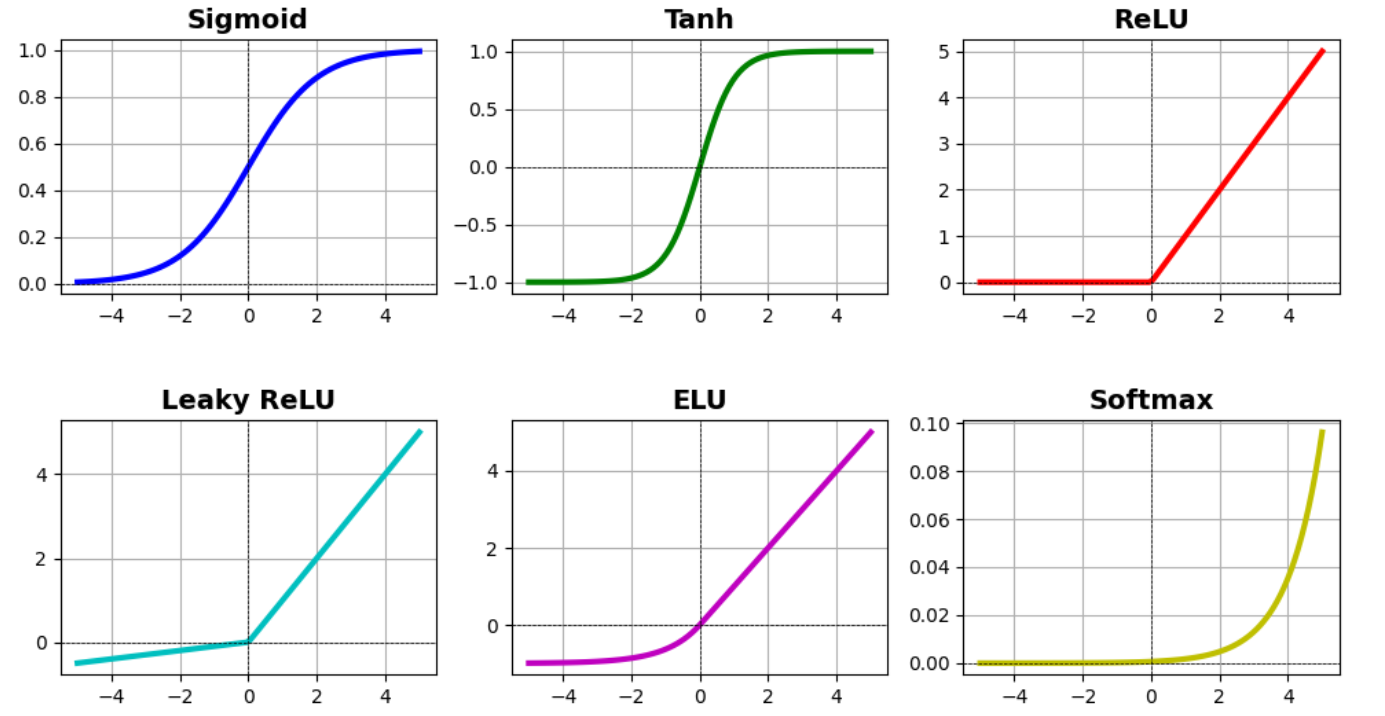}
\caption{Most commonly used Activation Functions.}
\label{activation_function}
\end{center}
\end{figure}

Feed-forward neural networks, also known as multilayer perceptrons, is the simplest neural network architecture widely used in various fields. The fundamental building block of a feed-forward neural network is the artificial neuron, which takes a set of inputs, performs a weighted sum of these inputs, and then applies activation function to produce an output. The mathematical representation of a feed forward neural network is given below.

\begin{equation*}
    Y = \sigma\bigg( \sum_{i=1}^N  W_i X_i + B\bigg)
\end{equation*}
where $X_i$ are the inputs, $W_i$ are the weights associated with each input and
$B$ is the bias term. The activation function is denoted by $\sigma$, which can be a sigmoid, hyperbolic tangent, rectified linear unit, or other nonlinear functions  \cite{ chigozie_nwankpa_c96fd363}. The architecture of a single layer neural network (also called perceptron) and multi-layer neural network are shown in Figure \ref{NN_figures}. The most commonly used activation functions are shown in Figure \ref{activation_function}. The neural network model is trained using back-propagation, which is a supervised learning method in which the weights and biases are updated to minimize the error between the predicted and the true outputs \cite{sornam2016survey}. If $\theta = \bigg[ W_1, W_2, ..., W_L, b_1, b_2, ...., b_L\bigg]$ represents the parameters of the model to be optimized during model training, where $W_i$ and $b_i$ are the weights and biases for $1 \leq i \leq L$, then the back-propagation steps can be described as follows.\\

\underline{\textbf{Back-propagation with Gradient Descent}}

\begin{enumerate}
    \item Initialize $i = 0$, \\
    $\theta_0 = \bigg[ W_1^0, W_2^0, ..., W_L^0, b_1^0, b_2^0, ...., b_L^0\bigg]$
    \item Define loss function = $L(\theta) =  \frac{1}{N}\sum_{i=1}^{n} (Y_i - f(X_i, \theta))^2$. Here $Y_i$ are actual outputs and $f(X_i, \theta)$ are model predicted outputs.
    \item while $|L(\theta_i)| > \epsilon$ do 
        \begin{itemize}
        \item[] (a) Calculate $\nabla \theta_i = \bigg[ \frac{\partial L(\theta)}{\partial W_{1, i}}, ...., \frac{\partial L(\theta)}{\partial W_{L, i}}, \frac{\partial L(\theta)}{\partial b_{1, i}}, ..., \frac{\partial L(\theta)}{\partial b_{L, i}}\bigg]$
        \item[] (b) Update $\theta_{i+1} = \theta_{i} - \eta \nabla \theta_i $, \hspace{.2cm} $\eta$ = learning rate
        \item[] (c) Increment $i = i+1$
    \end{itemize}
\end{enumerate}

Similar to the above machine learning models,  we considered two cases for the artificial neural network (ANN) model. In the first case, we used $70\%$ of the data for training, $10\%$ for validation, and the remaining $20\%$ for testing, and in the second case, the data split was  $80\%$, $10\%$, and $10\%$.  We used the relu activation function in the hidden layers and trained the model for 200 epochs with $0.001$ as the learning rate. The binary cross-entropy loss function (defined in equation \ref{lossentropy}) was used for the loss calculation.  The performance of the model is given in Table \ref{FNNresults}. We find that neural network performs extremely well in classifying cancer and non-cancer cases with an accuracy greater than $99 \%$. Comparing the performance with the machine learning models (Table \ref{performance_comparison_combined}), we find that the 2 hidden layer ANN model with relu activation function gives the highest accuracy, precision, recall, and F1 score.
\begin{equation}
\begin{split}
\text{Loss}(y, \hat{y}) =
&- \frac{1}{N} \sum_{i=1}^N \bigg[ y_i \log(\sigma(\hat{y}_i)) + (1 - y_i) \log(1 - \sigma(\hat{y}_i)) \bigg]
    \end{split}
    \label{lossentropy}
\end{equation}
where $y_i $ is the true label (0 or 1), $\hat{y}_i $ is the raw output (logit) from the model. $\sigma(\hat{y}_i) = \frac{1}{1 + e^{-\hat{y}_i}}$  is the sigmoid function that converts logits into probabilities.  The total number of samples in the batch is given by N.
\end{multicols}

\begin{multicols}{2}
\subsection{\textbf{Explanation of Model Prediction using SHAP and LIME}}
\begin{figure}[htbp!]
\begin{center}
\includegraphics[width=7.0in, height= 2.4in, angle=0]{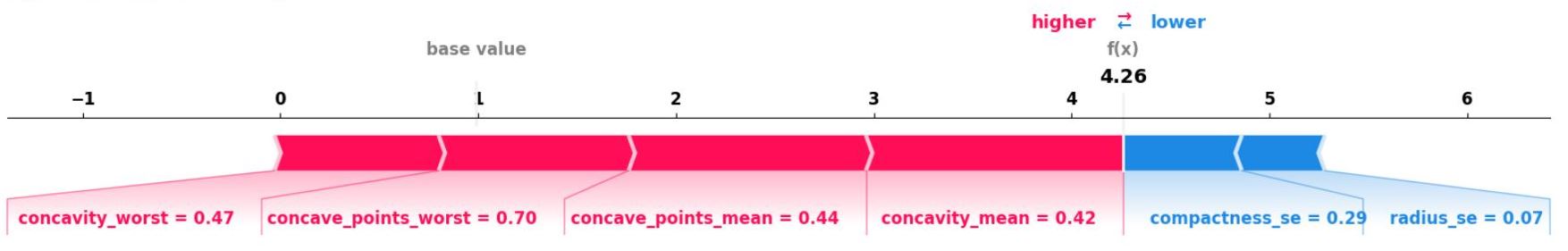}
\caption{Force plot: local interpretability of model prediction}
\label{forceplot}
\end{center}
\end{figure}

\begin{figure}[htbp!]
\begin{center}
\includegraphics[width=3.8in, height= 3.6in, angle=0]{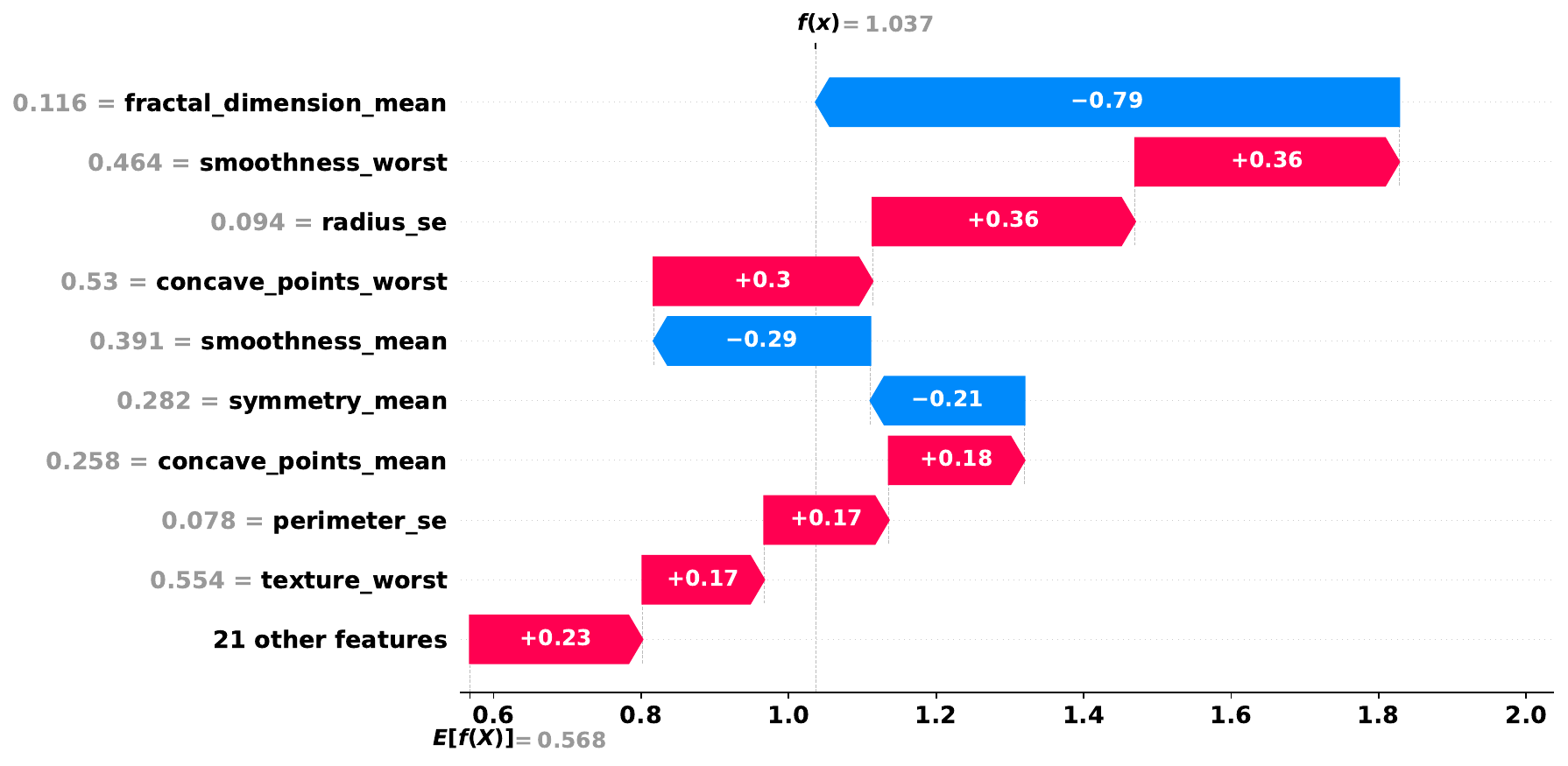}
\includegraphics[width=3.8in, height= 3.6in, angle=0]{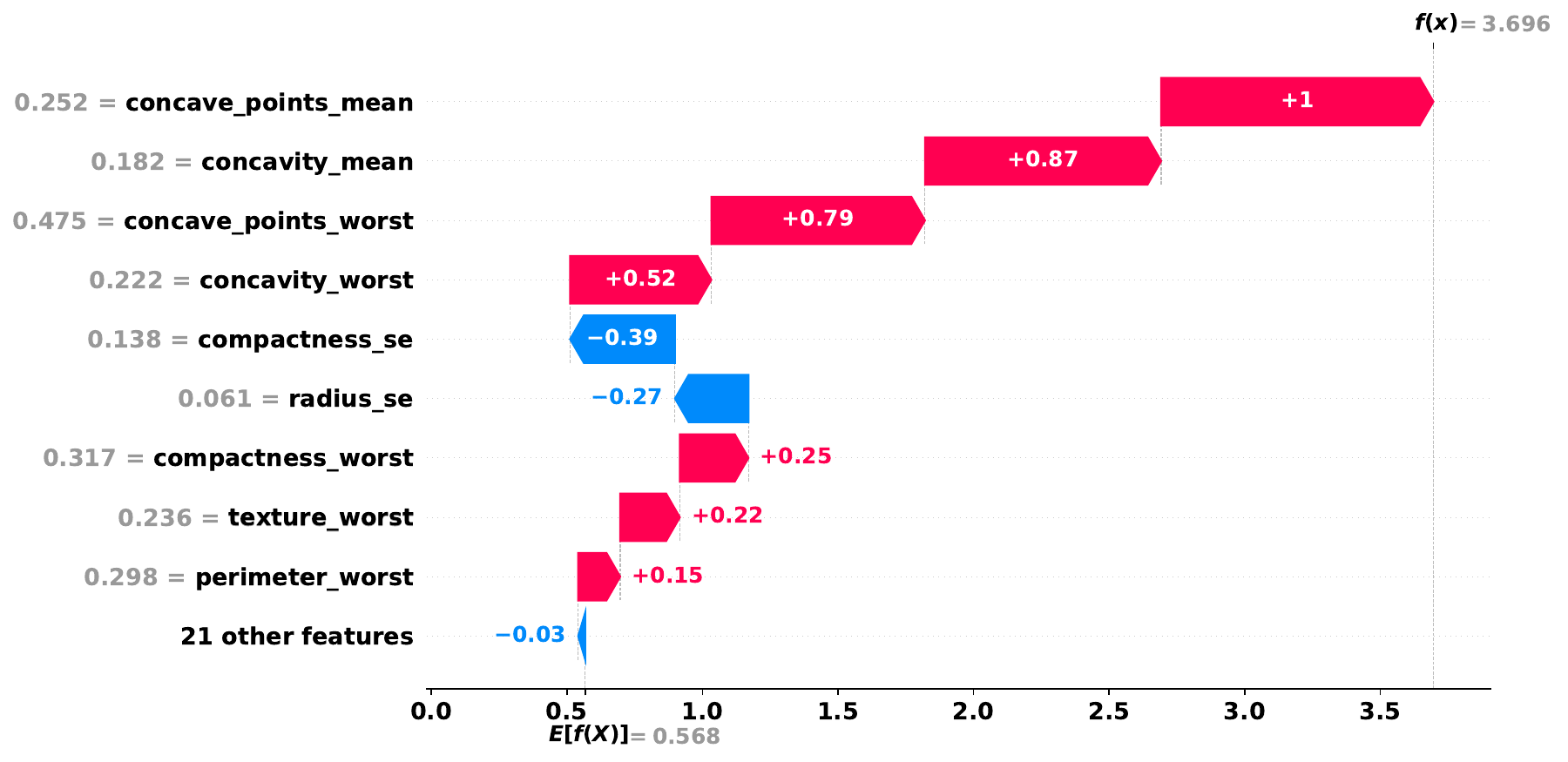}
\caption{Local Explanation: Waterfall plots for two test instances using Shapley values for neural network prediction.}
\label{waterfall}
\end{center}
\end{figure}

\begin{figure}[htbp!]
\begin{center}
\includegraphics[width=6.8in, height= 3.0in, angle=0]{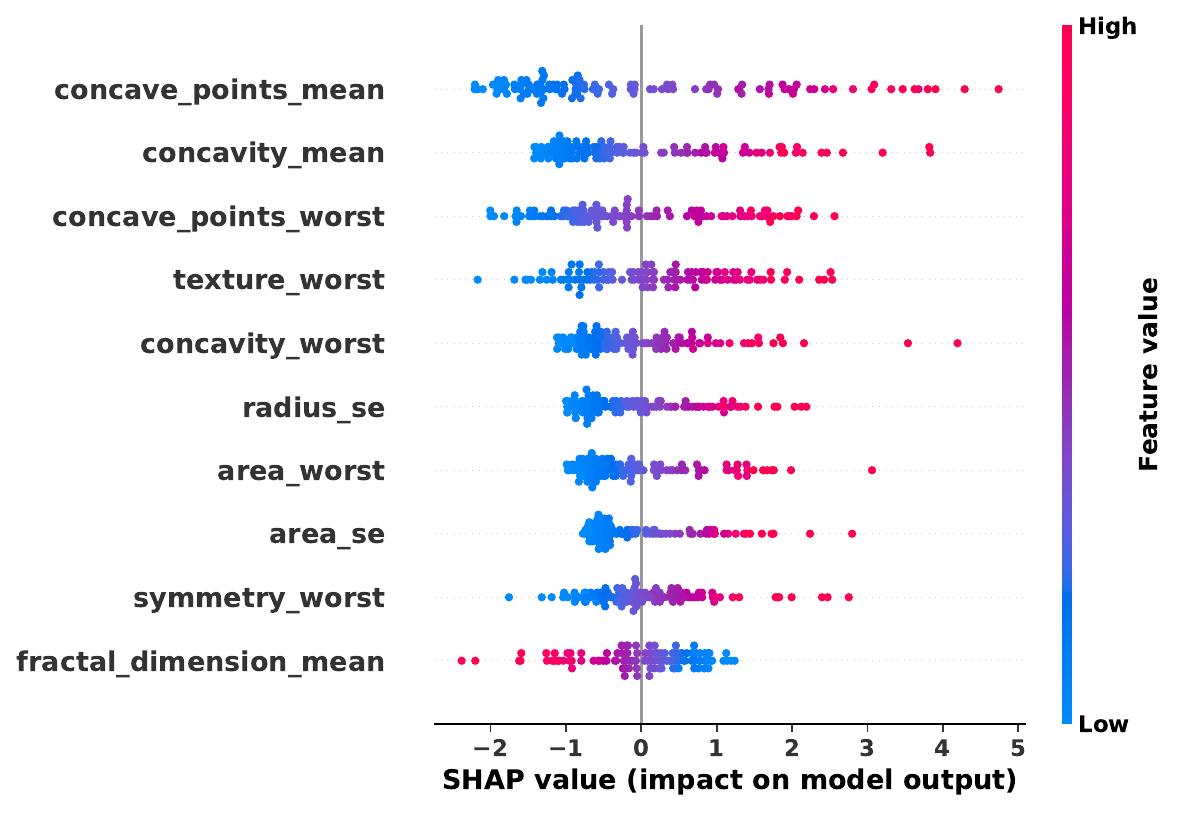}
\caption{Global Explanation: The SHAP summary plot shows the top 10  features that contribute to the neural network prediction.  We find that concave$\_$points$\_$mean feature has the highest contribution, followed by concavity mean. This means, the larger
the value of the concave points mean, the higher the chance
of the model predicting the case as malignant.}
\label{shap1}
\end{center}
\end{figure}

\begin{figure}[htbp!]
\begin{center}
\includegraphics[width=6.8in, height= 3.0in, angle=0]{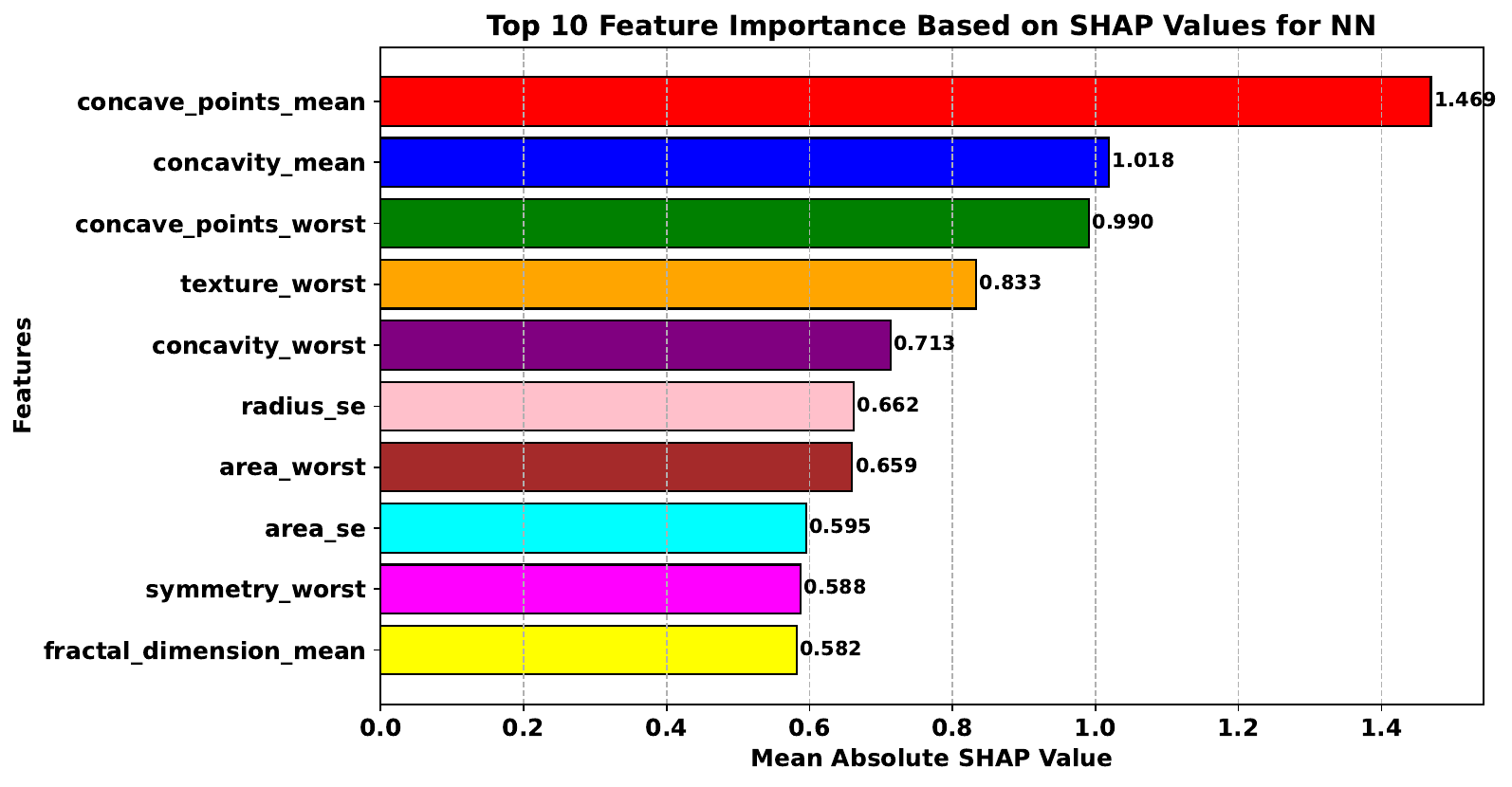}
\caption{Feature importance based on mean absolute SHAP-value for neural network prediction.}
\label{shap2}
\end{center}
\end{figure}

In this section, we want to understand the mechanism of the prediction by the ANN model. We use SHAP and LIME  methods for simplifying and explaining the black box character of the model. The shapely additive explanations(SHAP) technique is based on a game-theoretic approach. It considers each feature of the data as a player and the model outcome as the payoff and finds the contribution of each features for a prediction \cite{dwivedi2023explainable}. SHAP calculates a score for each feature which represents its weight to the model output. The local interpretable model-agnostic explanations(LIME), on the other hand, approximates any black box model with a local, interpretable model and explains each model prediction \cite{salih2024perspective}.  The SHAP force plot explaining a single prediction is shown in Figure \ref{forceplot}. The red bars represent the features that increase the predicted value, contributing toward a malignant diagnosis whereas the blue bars represent features that reduce the predicted value, pulling it back toward a benign diagnosis. The waterfall plots for the neural network predictions for two test instances are shown in Figure \ref{waterfall}.  The red parts of the plot show the features pushing the prediction higher from the base value (the average model output), i.e. these features contribute to predicting the instance as malignant, while the blue parts represent the features contributing to predicting the case as benign. The SHAP summary plot, and a mean absolute SHAP value plot are shown in Figures \ref{shap1}, \ref{shap2} for all the instances of the test data. This plots provides the global explanations of the model predictions.  The summary plot illustrates the relationship between feature values and their impact on the prediction. The ordering of the features are done according to their importance. Each point on the summary plot is a Shapley value for a feature at a particular instance. The color represents the value of the feature from low (blue) to high (red). For better visualization, the overlapping Shapley values are jittered along the y-axis. From these figures, we find that concave$\_$points$\_$mean feature has the highest contribution, followed by concavity$\_$mean. This means, that when the value of concave$\_$points$\_$mean is large, it positively affects the classification task. That is, the larger the value of the concave$\_$points$\_$mean, the higher the chance of the model predicting the case as malignant. In the case of fractal$\_$dimension$\_$mean, lower feature values is found to positively affects the classification task. \\

\begin{figure}[htbp!]
\begin{center}
\includegraphics[width=3.8in, height= 3.1in, angle=0]{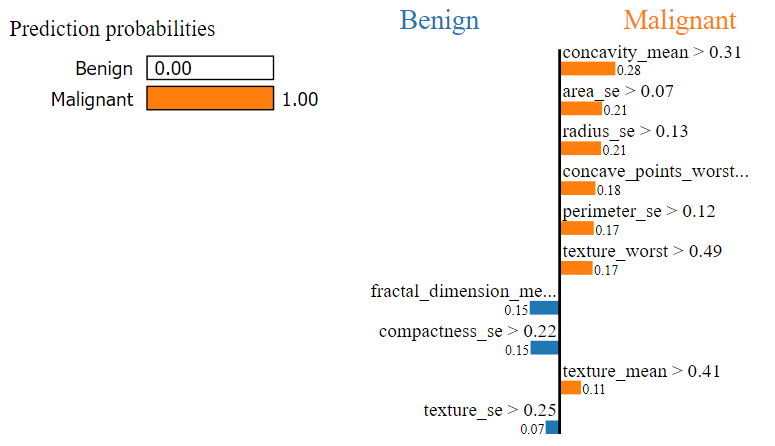}
\includegraphics[width=2.5in, height= 3.1in, angle=0]{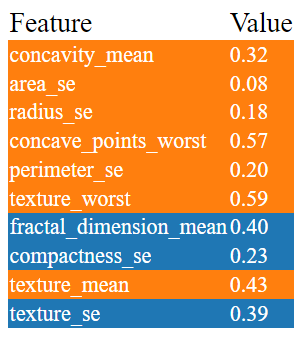}
\caption{LIME plot explaining artificial neural network prediction of malignant case with $100\%$ probability.  Top 10 features contributing for the prediction is illustrated. For this particular case, concavity$\_$mean is found be contribute highest for predicting the case as malignant. The model predicts the case as malignant with 100$\%$ probability.}
\label{lime2}
\end{center}
\end{figure}

\begin{figure}[htbp!]
\begin{center}
\includegraphics[width=3.8in, height= 3.0in, angle=0]{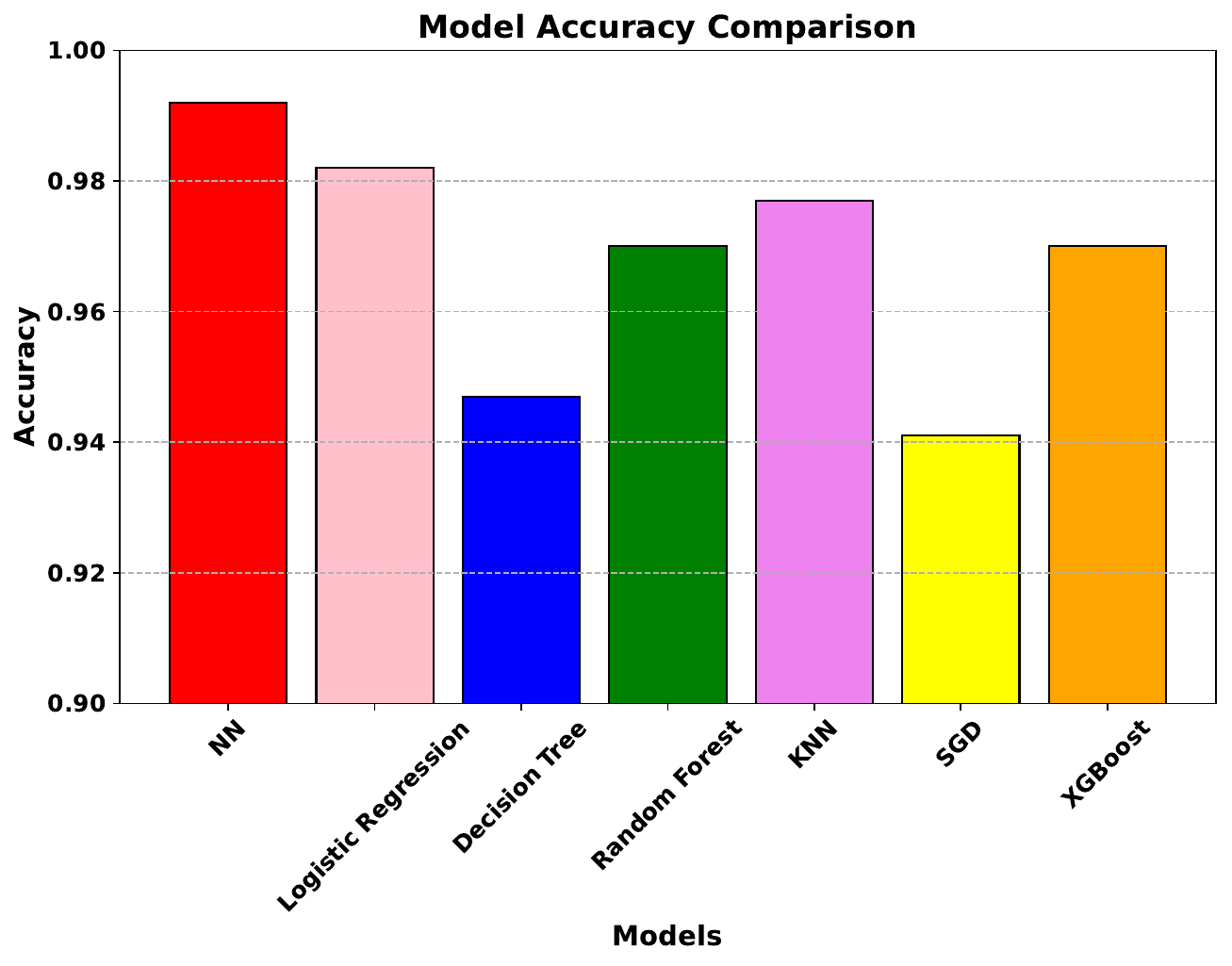}
\includegraphics[width=3.8in, height= 3.0in, angle=0]{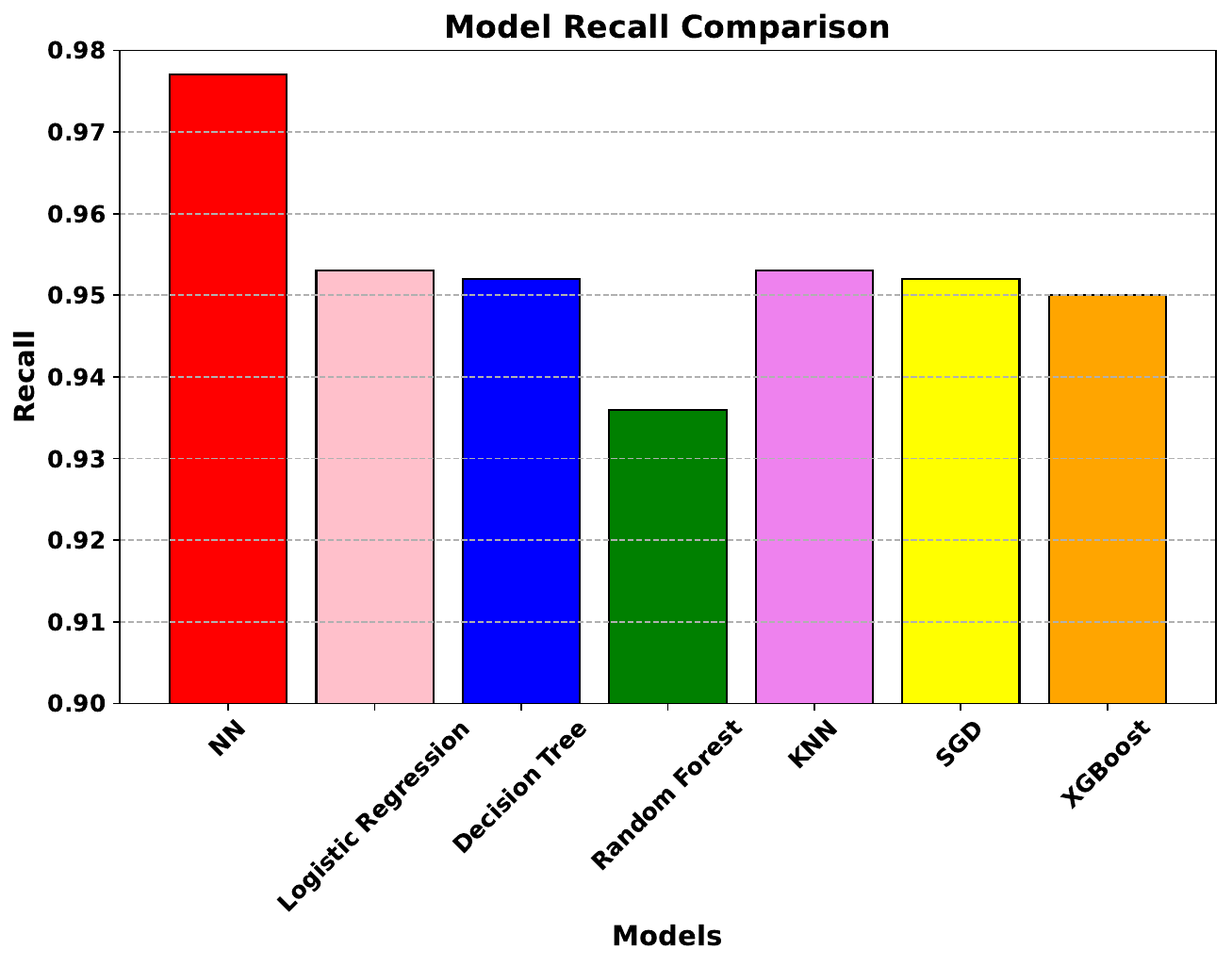}
\caption{Comparing Accuracy and Recall for Various Models. }
\label{accuracycompare}
\end{center}
\end{figure}

Figures \ref{lime2} illustrate the explanations generated by the LIME technique, for predictions made by neural network for a particular instance of the test dataset. The LIME technique generates prediction probabilities for benign and malignant classes with explanations, describing how the prediction is derived. The left part of the plot shows the probability that the instance of the data is classified as benign or malignant. The  middle part of the plot shows the weight or the coefficient value, of each feature
in the local linear model. The contribution of the top 10 features are illustrated in Figure \ref{lime2}, where the blue-colored features contribute to predicting the case as benign, while the yellow-colored features support malignancy. The actual feature values are shown on the right. From the figure, we see that the model predicts the given instance as malignant with 100 $\%$ accuracy, and concavity$\_$mean is found to contribute the maximum for the prediction.  In Figure \ref{accuracycompare}, bar plots are shown comparing the accuracy and recall of machine learning and ANN models. In Table \ref{paper_comparison}, we compare our results with the other state of art published works. In \cite{khater2023explainable}, an accuracy of $98.6\% $ is achieved using the artificial neural network. The authors also uses techniques such as permutation importance, PDP, and SHAP to interpret the model's predictions. In \cite{mohi2023xml}, LightGBM method is found to achieve an accuracy of $99\%$ in classifying malignant and benign cancer. The results of our proposed neural network model are comparable and better to those reported in the literature.
\end{multicols}

\begin{table}[htbp]
\centering
\captionof{table}{Performance Comparison of Classification Models}
\resizebox{0.5\linewidth}{!}{
\begin{tabular}{|c|c|c|c|c|c|}
\hline \hline
\textbf{Model} & \textbf{Data Split} & \textbf{Accuracy} & \textbf{Precision} & \textbf{Recall} & \textbf{F1 Score} \\
\hline
KNN & 7:3 & 0.977 & 0.984 & 0.952 & 0.968 \\
\cline{2-6}
& 8:2 & 0.974 & 0.976 & 0.953 & 0.965 \\
\hline
Logistic Regression & 7:3 & 0.971 & 1 & 0.936 & 0.967 \\
\cline{2-6}
& 8:2 & \textbf{\textcolor{red}{0.982}} & \textbf{\textcolor{red}{1}} & \textbf{\textcolor{red}{0.953}} & \textbf{\textcolor{red}{0.976}} \\
\hline
Decision Tree & 7:3 & 0.941 & 0.895 & 0.952 & 0.920\\
\cline{2-6}
& 8:2 & 0.947 & 0.930 & 0.930 & 0.930 \\
\hline
Random Forest & 7:3 & 0.970 & 0.983 & 0.936 & 0.960 \\
\cline{2-6}
& 8:2 & 0.956 & 0.957 & 0.930 & 0.941 \\
\hline
SGD & {7:3} & {0.941} & {0.90} & {0.952} & {0.925} \\
\cline{2-6}
& {8:2} & {0.932} & {0.920} & {0.942} & {0.930} \\
\hline
XGBoost & 7:3 & 0.970 & 0.953 & 0.950 & 0.951\\
\cline{2-6}
& 8:2 & 0.956 & 0.952 & 0.930 & 0.941 \\
\hline \hline

\end{tabular}
\label{performance_comparison_combined}
}
\end{table}

\begin{table}[htbp!]
\centering
\captionof{table}{Performance of NN model on Breast Cancer Diagnosis for Different Data Split Ratios (Train: Test: Validation)}
\begin{tabular}{|c|c|c|c|c|c|c|}
\hline \hline
\textbf{Data Split Ratio} & \textbf{Hidden Layers} &  \textbf{Accuracy} & \textbf{Precision} & \textbf{Recall} & \textbf{F1 Score} & \textbf{Train Accuracy}  \\
\hline\hline
\multirow{2}{*}{$7:2:1$} & 2 & \textbf{\textcolor{red}{0.992}} & \textbf{\textcolor{red}{1}}
 & \textbf{\textcolor{red}{0.977}} &  \textbf{\textcolor{red}{0.988}} & \textbf{\textcolor{red}{0.985}} \\
 \cline{2-7}
 & 3 & {0.980} & {0.976}
 & 0.953 &  0.964 & 0.983 \\
\hline
\multirow{2}{*}{$8:1:1$} & 2 & {{1}} & {{1}}
 & {{0.99}} &  {{0.995}} & {{0.985}} \\
 \cline{2-7}
 & 3 & {0.982} & {0.954}
 & 0.99 &  0.976 & 0.984 \\
\hline \hline
\end{tabular}
\label{FNNresults}
\end{table}

\begin{table}[ht!]
    \caption{Comparison With Published Works}
    \centering 
    \begin{tabular}{|c|c|c|c|c|c|} 
        \hline\hline
        
        \textbf{References} &  \textbf{Models used} &   \textbf{Best Model} &  \textbf{Accuracy} &  \textbf{Precision} &  \textbf{XAI 
        Technique} \\  
         \hline\hline
        \cite{chen2023classification} & XGBoost, RF, LR, KNN & XGBoost & 0.97 & 0.96 & Not used\\
        \hline\hline
        \cite{monirujjaman2022retracted} & DT, RF, LR, KNN & LR & 0.98 & 0.98 & Not used\\
     \hline\hline
        \cite{bhardwaj2022tree} & MLP, Genetic Programming (GP),  RF,  KNN & RF & 0.96 & 0.95 & Not used\\
        \hline\hline
        \cite{ahmed2020analysis} &Naive Bayes, MLP, J48, RF & Naive Bayes & 0.97 & 0.97 & Not used\\
        \hline\hline
        \cite{khater2023explainable} & KNN, SVM, XGBoost, RF, ANN & ANN & 0.986 & 0.944 & PDP\\
        & & & & & SHAP\\
        \hline\hline
   \cite{mohi2023xml} & LightGBM, GBM, XGBoost & LightGBM & 0.990 & 0.987 & SHAP \\
        \hline\hline
        Our study & DT, RF, SGD, XGBoost, LR, KNN, ANN  & ANN & 0.992 & 1.0 & SHAP and LIME \\
         \hline\hline
    \end{tabular}
    \label{paper_comparison}
\end{table}


\begin{multicols}{2}
    
\section{Discussion and Conclusion}
Cancer is one of the deadliest diseases of all time, with almost no cure if detected later.  Among various types of cancer, breast cancer is the most prevalent and a leading cause of female mortality in developing countries. Early detection or detection of the disease at an early stage is the only solution and is the key to successful treatment. The traditional approaches to early cancer detection include physical examination, imaging-based techniques, biopsy, and molecular tests, but these techniques come with several limitations and challenges. Imaging-based methods such as mammography sometimes fail to identify early-stage tumors, leading to false negative cases. Access to screening facilities could be limited because of various factors such as socioeconomic status, geographic location, healthcare disparities, etc. Another serious concern for cancer diagnosis and treatment is the financial challenge. The cost involved in precise diagnosis and treatment of cancer is excessively high. \\ 

AI models have become powerful tools for the early detection and diagnosis of various types of cancer.  In this study, we propose a deep neural network based models for early cancer detection.  We train the neural network model with Relu activation function for 200 epochs using 30 neurons in each layers, the Adam optimizer with a learning rate of 0.001, and the binary cross-entropy loss function for loss computation. The model’s classifying performance was evaluated using accuracy, precision, recall, and F1 score, which were found to be $0.992, 1.000, 0.977,$ and $0.988$ respectively. In addition, to highlight the efficacy of the proposed model, we compare the performance of the neural network model with other machine learning models such as logistic regression, decision tree, random forest, SGD, KNN, and XGBoost. Among all these models, the neural network model with 2 hidden layers and the ReLU activation function showed the highest values of all metrics used for the evaluation. To trust and deploy this model, it is important for the medical community to understand the reasons or logic behind the model prediction.  Therefore, to interpret the predictions, explainable machine learning techniques such as SHAP and LIME were used. The SHAP summary plots, force plots, waterfall plots and LIME plots are generated to understand the mechanisms of model prediction and assess the contributions of the data features to the decision-making process.  The concave points feature of cell nuclei is found to be the most influential feature positively impacting the classification task. Higher values of this feature are associated with an increased probability of the model classifying the case as malignant. This insight can be very helpful in improving the diagnosis and treatment of breast cancer by highlighting the key characteristics of breast tumors. In this study, the classification of breast cancer patients is done based on the features computed from fine needle aspiration of breast masses. Additional data sources such as clinical or genetic information can be extremely helpful, to build a robust model and make predictions.  Our future work will consider clinical and genetic information of the patients to build robust prediction models that integrate explainable AI techniques.\\

\noindent
\textbf{CRediT authorship contribution statement}\\
\noindent
BC developed the models, conducted the analysis, and wrote the article.
BVRK contributed in conceptual discussion, organization, supervision, and corrections.\\

\noindent
 \textbf{Conflict of Interest:} Authors declare no conflict of interest\\
 
\noindent
\textbf{Funding:} Not Applicable\\

\noindent
 \textbf{Data Availability:} Not Applicable \\

    \end{multicols}
\end{document}